\documentclass[10pt,a4paper,oneside]{article}
\usepackage{amssymb}
\usepackage{amsmath}
\usepackage{graphicx}
\usepackage{theorem}
\usepackage{setspace}
\usepackage{appendix}

\usepackage{booktabs}
\usepackage{longtable}
\usepackage{rotating}
\usepackage{multirow}
\usepackage{threeparttable}
\usepackage[font=small,labelfont=bf]{caption}
\usepackage{enumerate}
\usepackage{url}
\usepackage{setspace}
\begin{document}
\title{{\LARGE \textbf{Bearing fault diagnosis based on spectrum images of vibration signals}}}
\author{Wei Li$^{1}$, Mingquan Qiu$^{1}$, Zhencai Zhu$^{1}$, Bo Wu$^{1}$, and Gongbo Zhou$^{1}$\\
\\
$^{1}$School of Mechatronic Engineering,\\
China University of Mining and Technology,\\
Xuzhou, 221116, P.R. China. Email: liwei\_cmee@163.com\\
\\
}

\date{}
\maketitle

\begin{abstract}
Bearing fault diagnosis has been a challenge in the monitoring activities of rotating machinery, and it's receiving more and more attention. The conventional fault diagnosis methods usually extract features from the waveforms or spectrums of vibration signals in order to realize fault classification. In this paper, a novel feature in the form of images is presented, namely the spectrum images of vibration signals. The spectrum images are simply obtained by doing fast Fourier transformation. Such images are processed with two-dimensional principal component analysis (2DPCA) to reduce the dimensions, and then a minimum distance method is applied to classify the faults of bearings. The effectiveness of the proposed method is verified with experimental data. \bigskip

\bigskip
\end{abstract}

\section{Introduction}
Bearings are the most used components in rotating machinery, and the bearing faults may usually result in a great breakdowns and eventually casualties \cite{loparo2000fault,liu2012application,jalan2009model}. According to the statistics, bearing failure is about 40\% of the total failures of the induction motor’s \cite{bianchini2011fault}, and is the top contributor of gearbox faults in wind turbines \cite{link2011gearbox}. Hence it is important to diagnose bearings.

Fault diagnosis of bearings is usually based on vibration signals, and a set of features are extracted in order to classify the faults \cite{randall2011rolling}. The features could be in time domain, frequency domain or time-frequency domain \cite{jardine2006review}, such as peak amplitude, root-mean-square amplitude, skewness, kurtosis, correlation dimension, fractal dimension, Fourier spectrum, cepstrum, envelope spectrum \cite{heng1998statistical,samuel2005review}. These features are generally in forms of scalar or vector. Indeed they are some specific descriptions of waveform in time domain or some parameters of spectrum in frequency domain. A single feature only describes one aspect of vibration signals. Therefore many works combine more than one feature to improve the performance of fault diagnosis. For example, Khelf et al \cite{khelf2013adaptive} carried on fault diagnosis for rotating machines with several selected relevant features by doing indicators ranking according to a filter evaluation. And many other artificial intelligence methods for fault diagnosis often made full use of multi-features in time domain, frequency domain and time-frequency domain, so as to improve the diagnostic performance \cite{chen2014fault,wu2009automotive,li2012mechanical,xian2009intelligent,Li2015,lei2011application}.

The main way of human to recognize different objects is the vision, and the simplest form of vision is the image. Rich information is included in the image. Time domain and frequency domain features represent some characteristics of vibration signals. While the image is a much comprehensive description of vibration signals, and it could give much more information about the bearings. The computer vision techniques were well developed and applied for image processing and recognition \cite{hong1991algebraic,lee1999face,sun2005new,toole2007face}. Recently they were also further applied in the field of fault diagnosis \cite{klein2014bearing,grif2014enhanced}. In \cite{klein2014bearing}, an object detection method was used to detect specific lines in the time-frequency image of bearing vibration signals. In \cite{grif2014enhanced}, image processing method was employed to enhance the fault features in spectrogram of aircraft engines.

In this paper, we propose a novel fault diagnosis method using the spectrum image of vibration signal as the feature. The spectrum images of normal bearings and faulty bearings are obtained based on the fast Fourier transformation (FFT) of vibration signals, where all images are of the same sizes in pixels. We use two-dimensional principal component analysis (2DPCA) to process the images in order to reduce their dimensions and obtain the so-called eigen images, and then we classify the bearing faults with the help of a minimum distance method.

The rest of this paper is organised as follows. Section 2 presents the fault diagnosis method based on spectrum images, including image creation, image processing and image recognition. Section 3 gives the experimental results. Finally, the conclusion is given in section 4.

\section{Fault diagnosis based on spectrum images}
\subsection{Image creation with FFT}
\begin{figure} [!hbp]
\setlength{\abovecaptionskip}{5pt}            
\setlength{\belowcaptionskip}{0pt}            
\centering
  \includegraphics[width=0.7\textwidth,height=0.5\textheight]{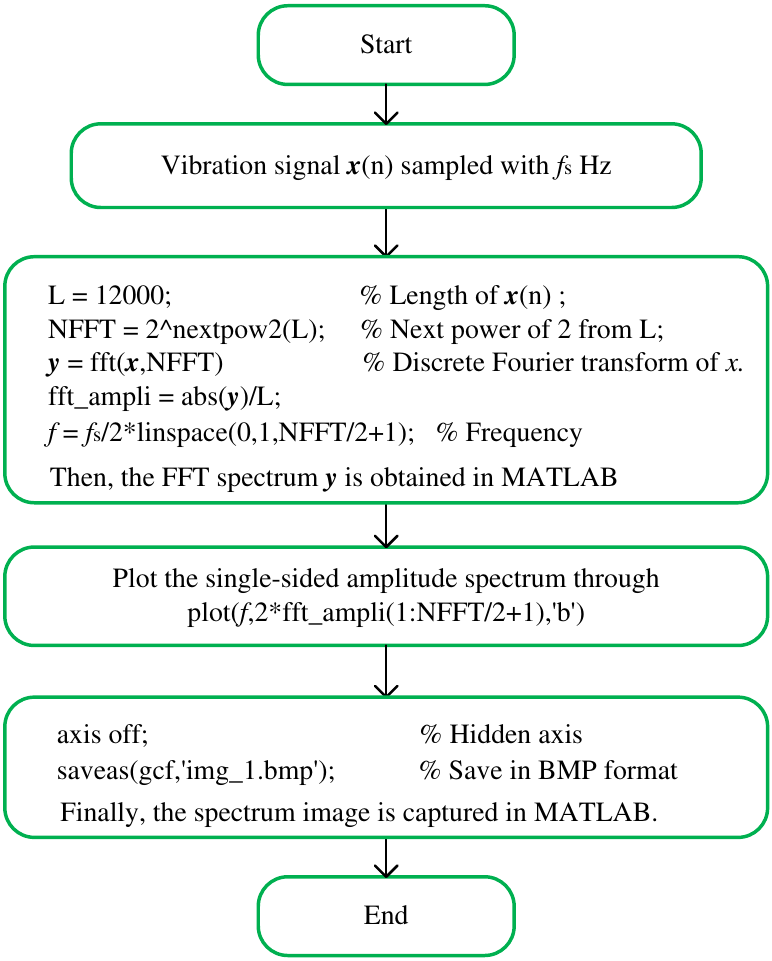}
  \caption{Flow chart of image creation in MATLAB.}\label{Image-Creation}
\end{figure}

There are many possible choices for the creation of vibration signal images. The images can be obtained in time domain, frequency domain, and time-frequency domain. In this study, we capture the FFT spectrums of vibration signals as images. The x-axis of the spectrum is frequency in the unit of Hertz, and the y-axis is the amplitude. For a given signal, the x-axis of the spectrum is determined by the sampling rate. When capturing the spectrum, its y-axis is auto-scaled. Then the parameter of the image is just the size (in pixels). With a larger image, more details of the spectrum can be depicted; while with a smaller image, some details may be lost. The flowchart of image creation in MATLAB is detailed illustrated in Fig. \ref{Image-Creation}.

By taking such images as the features of vibration signals, we actually make use of all information contained in the spectrum, i.e. characteristic frequencies and their harmonics of bearings, the geometrical structure of the spectrum, the peak amplitude of the spectrum and so on. On one side the images provide much more useful knowledge of bearings, and on the other side we can also take advantage of the well developed computer vision technique to realize fault diagnosis. In next the so-called 2DPCA is applied to process the obtained spectrum images, such that the low dimensional features of the images can be obtained.

\subsection{Image processing}
Similar with the principle of the conventional principal component analysis (PCA), 2DPCA is also a feature projection method, using which we can extract the intrinsic information of images with a direct operation on matrix. The projection process of 2DPCA can be concluded as follows \cite{yang2004two}.
\begin{itemize}
  \item \emph{Step} $1 \colon$ Suppose that there are \emph{M} training image samples, and the \emph{j}th training image(with $w \times h$ pixels) is denoted by an $w \times h$ matrix $\emph{A}_{j}, j=1,2,\ldots,M$. The average image of all the training image samples is denoted by $\overline{A}$. Then the global image scatter matrix \emph{G} can be evaluated by
      \begin{equation}\label{2dpcaeq1}
        G = \frac{1}{M}\sum_{j=1}^{N}(\emph{A}_{j}-\overline{A})^T(\emph{A}_{j}-\overline{A}) \in \mathbf{R}^{h \times h}
      \end{equation}
      where $(\bullet)^T$ represents the transpose of matrix $(\bullet)$.   
  \item \emph{Step} $2 \colon$ In order to obtain the basis vectors, it is necessary to find the eigenvectors $u_{k}$ and eigenvalues $\lambda_{k}$ of the global image scatter matrix \emph{G} through solving the following equation:
      \begin{equation}\label{2dpcaeq2}
        Gu = \lambda u
      \end{equation}
      where $\lambda_{k}, k=1,2, \ldots ,h$, is the eigenvalues, and $u=[u_{1}, u_{2}, \ldots, u_{h}]$ is the corresponding eigenvectors of \emph{G}. Also for reducing dimensions and decreasing computational expense in next, we normalise and sort the eigenvectors $u$ decreasing order according to the corresponding eigenvalues $\lambda_{k}$. Then the $u=[u_{max1},u_{max2}, \ldots ,u_{maxh}]$ and the corresponding eigenvalues $\lambda=[\lambda_{max1}, \lambda_{max2}, \ldots, \lambda_{maxh}]$ can be obtained. Here $\lambda_{k}$ satisfies the following constraint:
      \[
      \lambda_{max1}>\lambda_{max2}>\ldots>\lambda_{maxh}
      \]
  \item \emph{Step} $3 \colon$ In order to obtain the optimal projection vectors, the first $d(d<h)$ largest eigenvectors are selected to form the projection basis as
      \begin{equation}\label{2dpcaeq3}
        U = [u_{max1} \quad u_{max2} \quad \cdots \quad u_{maxd}]
      \end{equation}

  \item \emph{Step} $4 \colon$ Feature extraction is implemented with the projection basis obtained in the previous step. For a given image sample $B$, which is also the same size of $w \times h$ pixels corresponding to $A_{j}$, let
      \begin{equation}\label{2dpcaeq4}
        Y_{k}=BU_{k}
      \end{equation}
      where, $U_{k}=u_{maxk}$, and $k=1,2, \ldots ,d$.
      Then the projected feature vectors, $Y_{1},Y_{2}, \ldots ,Y_{d}$, can be obtained, which are called the principal components of the image sample $B$. At last we can obtain the eigen image of $B$ in the form of
      \begin{equation}\label{2dpcaeq5}
        E = [Y_{1},Y_{2}, \ldots ,Y_{d}] \in \mathbf{R}^{w \times d}
      \end{equation}
\end{itemize}

\subsection{Image recognition}
In order to classify the bearing faults, spectrum images of different faulty bearings must be recognized. Firstly some training image samples are processed through 2DPCA to obtain the corresponding eigen images of vibration signals with different faults. Then a nearest neighbor classification method is utilized for the classification of testing spectrum images.

Suppose that the $i$th projection feature matrix of the $M$ training image samples is $E_{i} = [Y_{1}^{(i)}, Y_{2}^{(i)}, \ldots, Y_{d}^{(i)}]$, where $i = 1, 2, \ldots, M$, and that of the $j$th testing image is $T_{j} = E_{j} = [Y_{1}^{(j)}, Y_{2}^{(j)}, \ldots, Y_{d}^{(j)}]$. Here we apply Euclidean distance \cite{wang2005euclidean} to measure the distance between $E_{i}$ and $T_{j}$ as follows
\begin{equation}\label{Euclidean_distance}
  d_{i}(E_{i}, T_{j}) = L_{p=2}(E_{i}, T_{j}) = \sum_{r=1}^{d} \|Y_{r}^{(i)}-Y_{r}^{(j)}\|_{2}
\end{equation}
where $\|Y_{r}^{(i)}-Y_{r}^{(j)}\|_{2}$ denotes the Euclidean distance between $Y_{r}^{(i)}$ and $Y_{r}^{(j)}$, and $Y_{r}^{(i)}$, $Y_{r}^{(j)}$ are the $r$th vector of $E_{i}$ and $T_{j}$.

Suppose $L = {s_{1}, s_{2},  \ldots, s_{N},}(N \leq M)$ is the category label set of the $M$ training samples. In order to classify the $j$th testing image, it’s necessary to find the subscript $\eta$, which satisfies
\begin{equation}\label{min_dist}
  d_{\eta} = min \ (d_{i})
\end{equation}
Then if the $\eta$th training image is assigned as $s_{k}(s_{k} \in L)$, the $j$th testing image is classified as $s_{k}$.

\subsection{Main procedure of the method}
The main procedure of fault diagnosis process based on spectrum images is summarized as follows.
\begin{itemize}
  \item \emph{Data Acquisition:} The spectrum images of vibration signals are firstly created through FFT as described in section 2.1. The image database can be constructed with these spectrum images.
  \item \emph{Eigen-images Extraction:} Once the image database is set up, the eigen images could be obtained through 2DPCA as given in section 2.2.
  \item \emph{Fault Classification:} A testing spectrum image can be classified by comparing with training spectrum images using the method given in section 2.3.
\end{itemize}

\begin{figure} [!hbp]
\setlength{\abovecaptionskip}{5pt}            
\setlength{\belowcaptionskip}{0pt}            
\centering
  \includegraphics[width=1.08\textwidth,height=0.7\textheight]{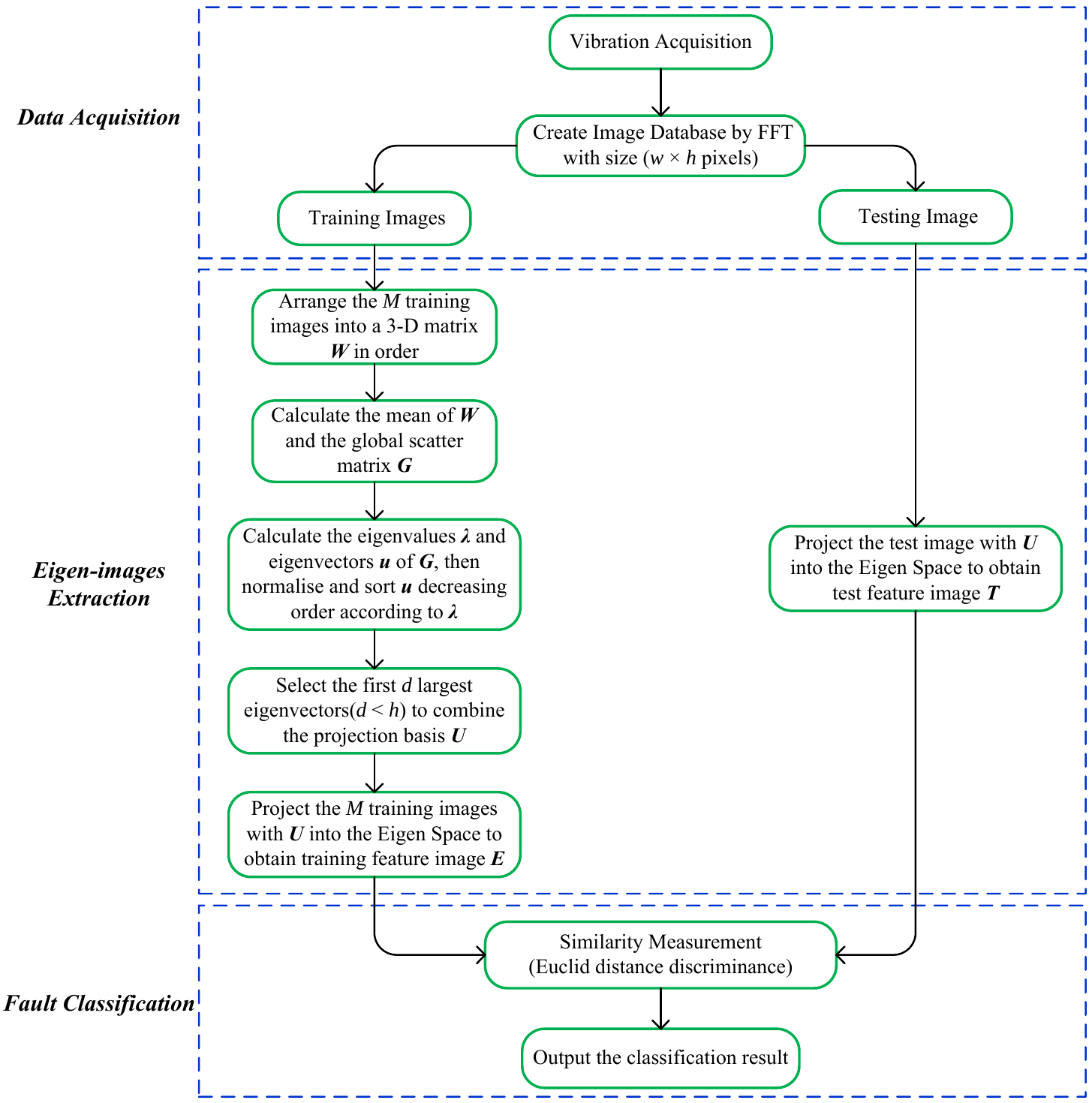}
  \caption{Flow chart of bearing fault diagnosis based on spectrum images.}\label{2DPCAdiagnosis-Green}
\end{figure}

The flow chart of bearing fault diagnosis based on spectrum images is shown in Fig. \ref{2DPCAdiagnosis-Green}.

\section{Experimental results}
In order to verify the effectiveness of the proposed fault diagnosis method, the vibration signals from the bearing data centre of Case Western Reserve University are used \cite{CWRUBDC-Bearings}. The test stand consists of a driving motor, a 2 hp motor for loading, a torque transducer/encoder, accelerometers and control electronics. The test bearings support the motor shaft. With the help of electrostatic discharge machining, inner-race faults (IF), outer-race faults (OF) and ball faults (BF) of different sizes (0.007in, 0.014in, 0.021in and 0.028in) are made. The vibration signals are collected using accelerometers attached to the housing with magnetic bases, and four load conditions with different rotating speeds were considered, i.e. Load0=0hp/1797rpm, Load1=1hp/1772rpm, Load2=2hp/1750rpm and Load3=3hp/1730rpm. The vibration signals of normal bearings (NO) under different load conditions were also collected.

Rotating machinery usually works under different loads and speeds. In practice it is common to obtain features of faults under a certain load condition. Hence the vibration signals with two fault sizes  (0.014in and 0.021in) under all four load conditions are used to demonstrate the effectiveness of the proposed method, where the training images of IF, OF and BF are created from one load condition (called training load condition) and the testing images are from all four load conditions (called testing load condition). Totally 8 different tests are carried out as shown in Table \ref{CWRU-faulty-bearings} to classify the faults of bearings.

The FFT spectrum of each vibration signal is computed by using 1024 sampling points. The y-axis of the spectrum is auto-scaled. Then the spectrum is captured as an image with the size of $420 \times 560$ pixels. Fig. \ref{Normal} - Fig. \ref{Outer-Fault} show the spectrum images of a normal bearing and faulty bearings with fault size being 0.014in. Four hundred spectrum images are generated for normal bearing (NO) and faulty bearings (IF, BF or OF) under each load condition. The training images of normal bearings and faulty bearings are selected randomly under the training load condition, and all 400 spectrum images under the each testing load condition are used for verification. Each test in Table \ref{CWRU-faulty-bearings} is performed for 20 times and the average classification rate is obtained.

\begin{table}[!hbp] \footnotesize
  \centering
  \renewcommand\arraystretch{1.4}               
  \renewcommand{\multirowsetup}{\centering}
  \setlength{\abovecaptionskip}{5pt}            
  \setlength{\belowcaptionskip}{0pt}            
  \caption{Description of the experiment setup.}\label{CWRU-faulty-bearings}
  \begin{tabular}{ccccc}
    \toprule[1pt]
    \# of test & Training & Testing & Fault type & Fault size  \\
    \hline
    1 & Load0 & Load0, Load1, Load2, Load3 & IF, BF, OF, NO & 0.014in \\
    2 & Load1 & Load0, Load1, Load2, Load3 & IF, BF, OF, NO & 0.014in \\
    3 & Load2 & Load0, Load1, Load2, Load3  & IF, BF, OF, NO & 0.014in \\
    4 & Load3 & Load0, Load1, Load2, Load3  & IF, BF, OF, NO & 0.014in \\
    5 & Load0 & Load0, Load1, Load2, Load3  & IF, BF, OF, NO & 0.021in \\
    6 & Load1 & Load0, Load1, Load2, Load3  & IF, BF, OF, NO & 0.021in \\
    7 & Load2 & Load0, Load1, Load2, Load3  & IF, BF, OF, NO & 0.021in \\
    8 & Load3 & Load0, Load1, Load2, Load3  & IF, BF, OF, NO & 0.021in \\
    \bottomrule[1pt]
  \end{tabular}
\end{table}

\begin{figure} [!hbp]
\setlength{\abovecaptionskip}{5pt}            
\setlength{\belowcaptionskip}{0pt}            
\centering
  \includegraphics[width=\textwidth,height=0.32\textheight]{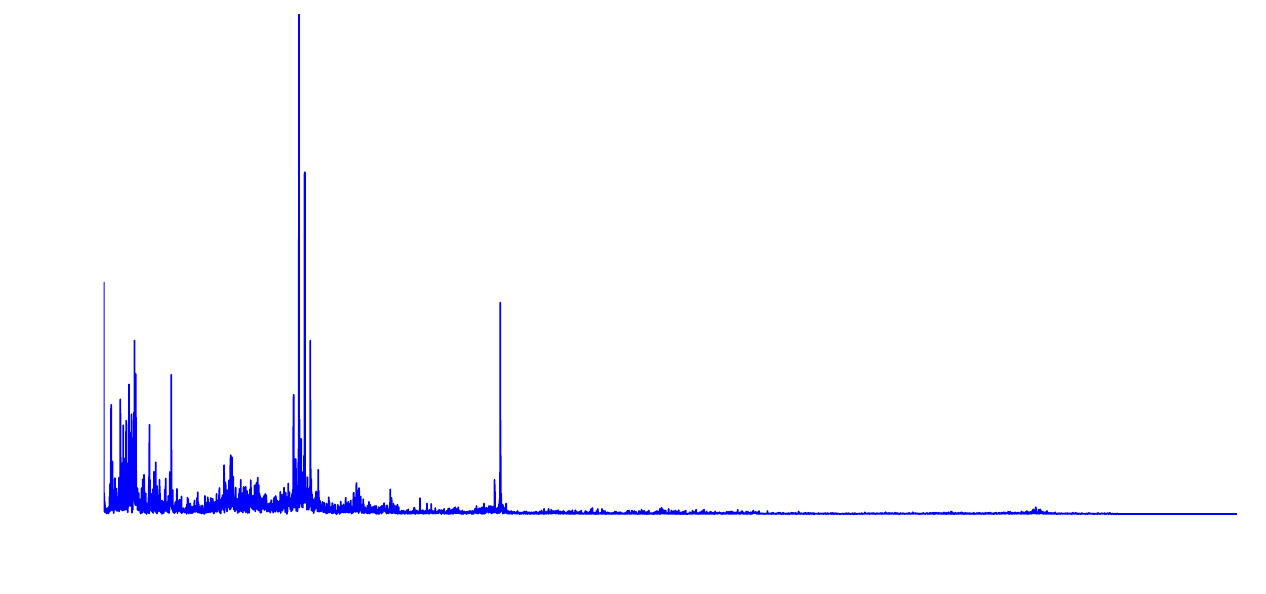}
  \caption{The FFT spectrum image of a normal signal.}\label{Normal}
\end{figure}

\begin{figure} [!hbp]
\setlength{\abovecaptionskip}{5pt}            
\setlength{\belowcaptionskip}{0pt}            
\centering
  \includegraphics[width=\textwidth,height=0.32\textheight]{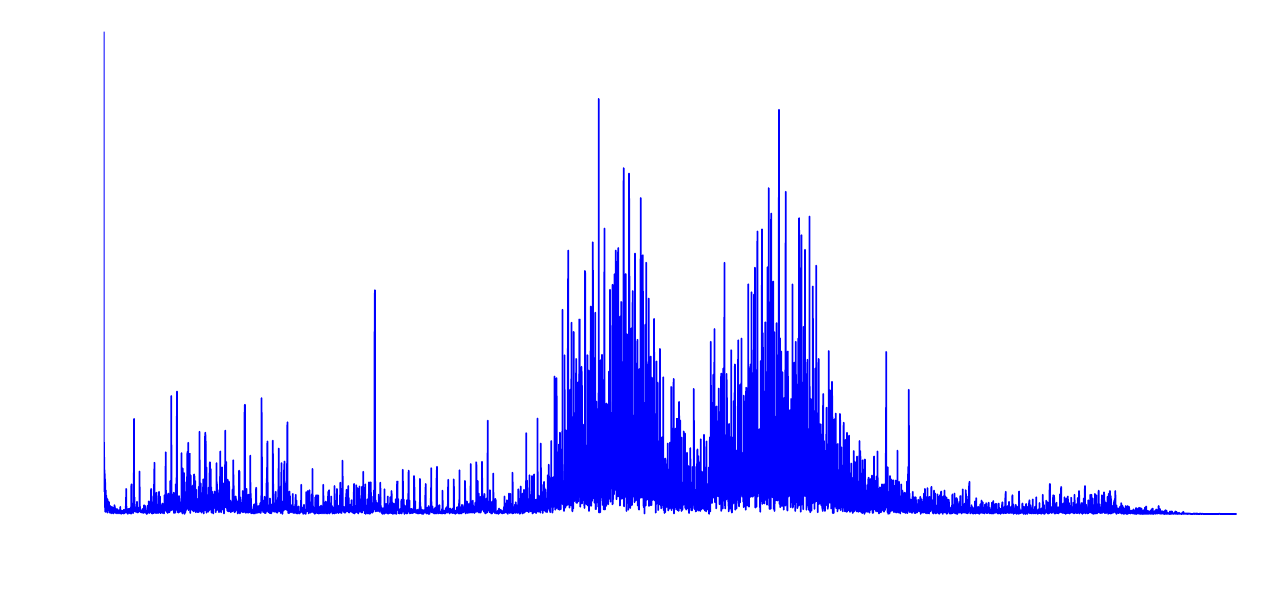}
  \caption{The FFT spectrum image of an inner-race fault signal.}\label{Inner-Fault}
\end{figure}

\begin{figure} [!hbp]
\setlength{\abovecaptionskip}{5pt}            
\setlength{\belowcaptionskip}{0pt}            
\centering
  \includegraphics[width=\textwidth,height=0.32\textheight]{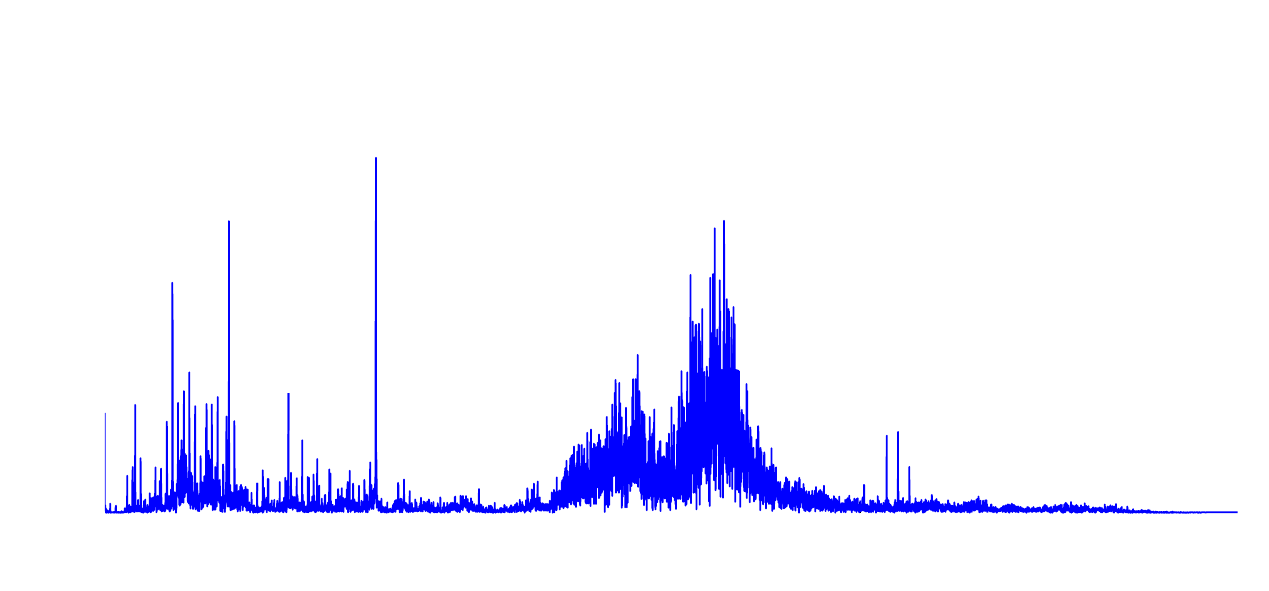}
  \caption{The FFT spectrum image of a ball fault signal.}\label{Ball-Fault}
\end{figure}

\begin{figure} [!hbp]
\setlength{\abovecaptionskip}{5pt}            
\setlength{\belowcaptionskip}{0pt}            
\centering
  \includegraphics[width=\textwidth,height=0.32\textheight]{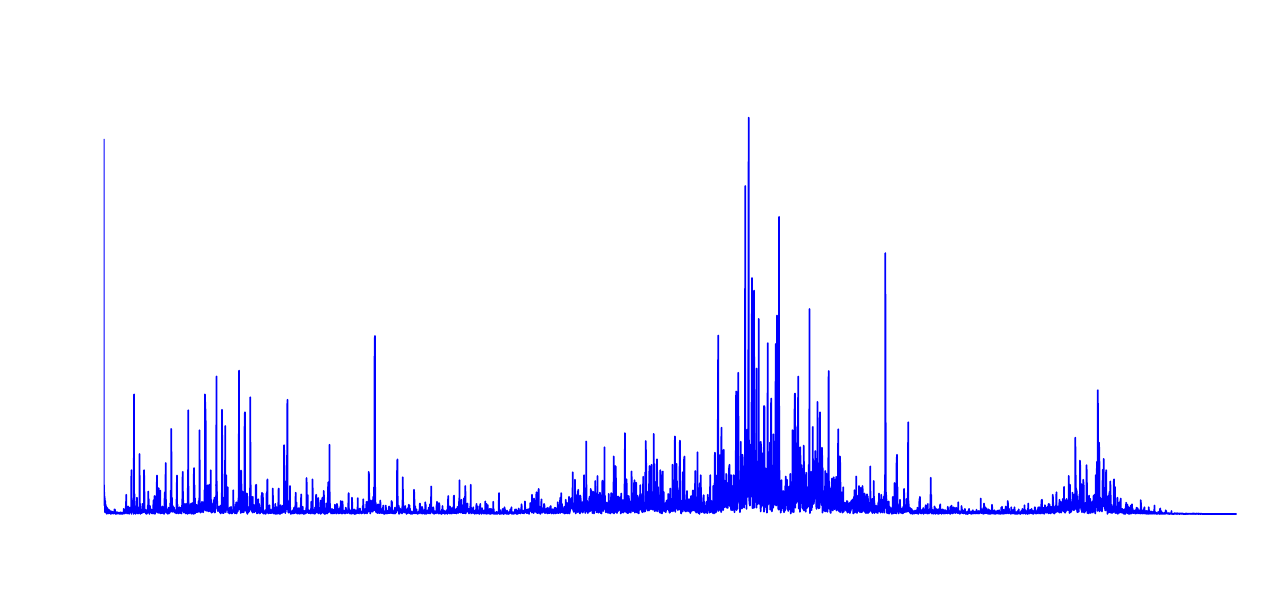}
  \caption{The FFT spectrum image of an outer-race fault signal}\label{Outer-Fault}
\end{figure}

In order to demonstrate the effectiveness of the proposed method, the spectrum images are processed through PCA and 2DPCA, and the classification results between them are compared.

\subsection{Results based on PCA}
The tests are firstly demonstrated with PCA. According to the feature dimensionality reduction criterion in \cite{abdi2010principal}, the so-called contribution of selected components with 20\%, 40\%, 60\% ,80\%, 90\% and 100\% are firstly performed to determine the reduced dimension. Thereafter dimension reduction with 90\% is designated in our research, as in this case the average classification rate is the highest. The experimental results is shown in Table \ref{Result-0.014-PCA} and Table \ref{Result-0.021-PCA}. 

\begin{table}[!hbp] \footnotesize
  \centering
  \renewcommand\arraystretch{1.1}                   
  \renewcommand{\multirowsetup}{\centering}
  \setlength{\abovecaptionskip}{5pt}                
  \setlength{\belowcaptionskip}{0pt}                
  \renewcommand{\thefootnote}{\fnsymbol{footnote}}  
  \caption{The classification rate based on PCA with fault size being 0.014.}\label{Result-0.014-PCA}
  \begin{threeparttable}
    \begin{tabular}{ccccccc}
      \toprule[1pt]
      \multirow{2}{1cm}{\# of test} & \multirow{2}{0.4cm}{$n$\tnote{a}}
      & \multicolumn{5}{c}{Testing data} \\\cline{3-7}
      &  &  \multirow{1}{*}{Test1(\%)}  & \multirow{1}{*}{Test2(\%)} & \multirow{1}{*}{Test3(\%)}
      & \multirow{1}{*}{Test4(\%)} & \multirow{1}{*}{$T$(s)\tnote{b}} \\
      \hline
      1 & $1$ & Load0(97.15)  & Load1(99.95)  & Load2(97.74) & Load3(93.41) & 66 \\
      $$ & $3$ & Load0(98.15)  & Load1(99.96)  & Load2(99.99)  & Load3(99.65) & 80 \\
      $$ & $5$ & Load0(99.99)  & Load1(100.00)  & Load2(100.00)  & Load3(100.00) & 95 \\
      $$ & $10$ & Load0(100.00)  & Load1(100.00)  & Load2(100.00) & Load3(100.00) & 145 \\
       $$ & $$ & $$ & $$ & $$ & $$ & $$ \\              

      2 & $1$ & Load0(83.80)  & Load1(97.15)  & Load2(100.00) & Load3(75.00) & 64 \\
      $$ & $3$ & Load0(84.60)  & Load1(100.00) & Load2(100.00)  & Load3(75.04) & 77 \\
      $$ & $5$ & Load0(86.40)  & Load1(100.00)  & Load2(100.00) & Load3(75.00) & 92 \\
      $$ & $10$ & Load0(86.04)  & Load1(100.00)  & Load2(100.00)  & Load3(75.00) & 138 \\
       $$ & $$ & $$ & $$ & $$ & $$ & $$ \\              

      3 & $1$ & Load0(84.92)  & Load1(99.76)  & Load2(97.15) & Load3(76.05) & 68 \\
      $$ & $3$ & Load0(86.17)  & Load1(100.00)  & Load2(100.00)  & Load3(77.03) & 80 \\
      $$ & $5$ & Load0(88.70)  & Load1(99.96)  & Load2(100.00)  & Load3(75.26) & 94 \\
      $$ & $10$ & Load0(89.45)  & Load1(100.00) & Load2(100.00)  & Load3(75.75) & 144 \\
      $$ & $$ & $$ & $$ & $$ & $$ & $$ \\              

      4 & $1$ & Load0(96.56)  & Load1(85.75)  & Load2(98.30)  & Load3(97.15) & 66 \\
      $$ & $3$ & Load0(96.61)  & Load1(86.81)  & Load2(96.26)  & Load3(100.00) & 81 \\
      $$ & $5$ & Load0(97.70) & Load1(84.76)  & Load2(97.30)  & Load3(100.00) & 97 \\
      $$ & $10$ & Load0(96.99)  & Load1(77.72)  & Load2(97.01)  & Load3(100.00) & 148 \\
      \bottomrule[1pt]
    \end{tabular}
    \begin{tablenotes}
      \footnotesize
      \item[a] $n$ is the number of training samples per class, and the same below.
      \item[b] $T$ is the total time consumption of the processing program from \emph{Eigen-images Extraction} to \emph{Fault Classification} for 20 times randomized tests, and the same below as well.
    \end{tablenotes}
  \end{threeparttable}
\end{table}

\begin{figure} [!hbp]
\setlength{\abovecaptionskip}{5pt}            
\setlength{\belowcaptionskip}{0pt}            
  \centering
  \includegraphics[width=\textwidth,height=0.32\textheight]{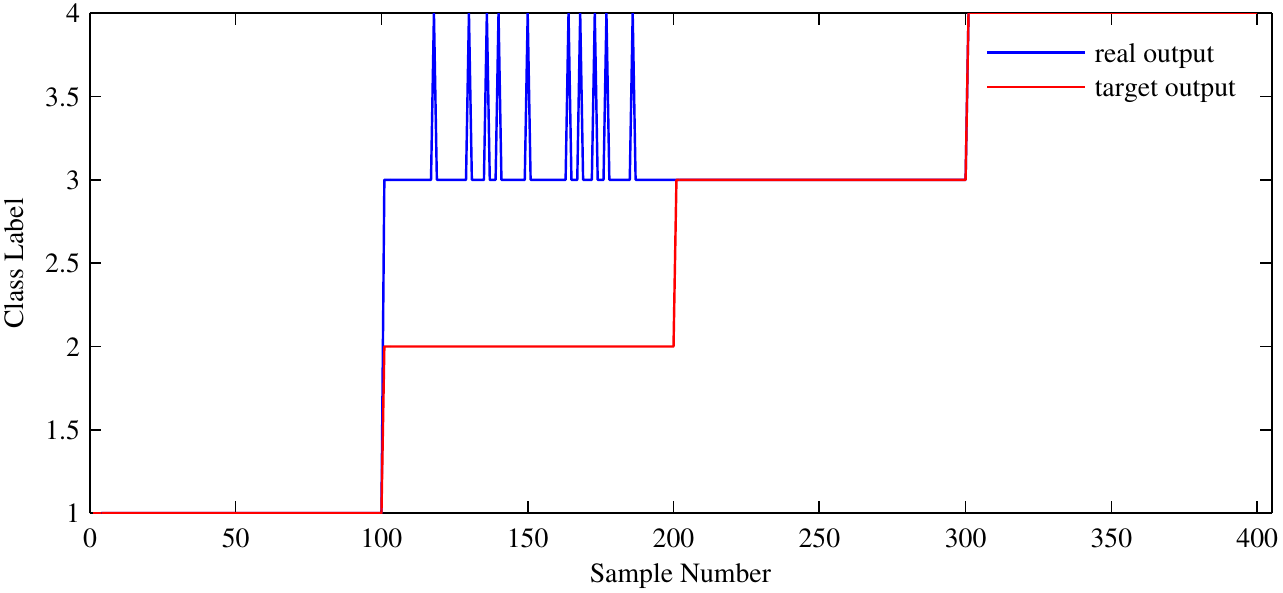}
  \caption{Actual output of Load1 for training and Load3 for testing with fault size being 0.014.}\label{pca-0.014-output}
\end{figure}

From Table \ref{Result-0.014-PCA} we can see that the classification rates are mostly larger than $90\%$, but the classification rates of several tests are relatively low. For example, we only obtain a classification rate about 75\% when using images under Load1 as training data and images under Load3 as testing data. The actual output with $n = 3$ in this case is shown as Fig. \ref{pca-0.014-output}. We can see intuitively that the IF (target output is: 2) images are classified incorrectly. Some testing images of IF are categorized into BF (target output is: 3), and the others are categorized into OF (target output is: 4) incorrectly. Through observing and analyzing the spectrum images of IF (Load3) and BF (Load1) carefully, it is plausible that the spectrum images of IF (Load3), BF (Load3) and BF (Load1) looked very similar, which resulted in the low classification rate.

Similar tests are carried out on the spectrum images with fault size being 0.021, and the classification results are presented in Table \ref{Result-0.021-PCA}. We can observe that the overall classification rate is higher than that of fault size being 0.014. It is worth mentioning that an acceptable classification rate can be achieved by using only single training image ($n=1$).

\begin{table} \footnotesize
  \centering
  \renewcommand\arraystretch{1.1}               
  \renewcommand{\multirowsetup}{\centering}
  \setlength{\abovecaptionskip}{5pt}            
  \setlength{\belowcaptionskip}{0pt}            
  \caption{The classification rate based on PCA with fault size being 0.021.}\label{Result-0.021-PCA}
     \begin{tabular}{ccccccc}
      \toprule[1pt]
      \multirow{2}{1cm}{\# of test} & \multirow{2}{0.4cm}{$n$\tnote{a}}
      & \multicolumn{5}{c}{Testing data} \\\cline{3-7}
      &  &  \multirow{1}{*}{Test1(\%)}  & \multirow{1}{*}{Test2(\%)} & \multirow{1}{*}{Test3(\%)}
      & \multirow{1}{*}{Test4(\%)} & \multirow{1}{*}{$T$(s)\tnote{b}} \\
      \hline
    5 & $1$ & Load0(97.65)  & Load1(99.13)  & Load2(92.94)  & Load3(87.22) & 65 \\
    $$ & $3$ & Load0(97.92)  & Load1(100.00)  & Load2(98.83)  & Load3(99.26) & 80 \\
    $$ & $5$ & Load0(99.75)  & Load1(98.67)  & Load2(96.89)  & Load3(97.28) & 90 \\
    $$ & $10$ & Load0(99.75)  & Load1(98.26)  & Load2(97.35)  & Load3(97.71) & 134 \\
     $$ & $$ & $$ & $$ & $$ & $$ & $$ \\              

    6 & $1$ & Load0(99.66)  & Load1(99.75)  & Load2(90.36)  & Load3(93.86) & 66 \\
    $$ & $3$ & Load0(99.75)  & Load1(100.00)  & Load2(92.28)  & Load3(97.06) & 78 \\
    $$ & $5$ & Load0(99.75)  & Load1(100.00) & Load2(93.99)  & Load3(97.69) & 92 \\
    $$ & $10$ & Load0(99.75) & Load1(100.00)  & Load2(94.74)  & Load3(96.66) & 138 \\
     $$ & $$ & $$ & $$ & $$ & $$ & $$ \\              

    7 & $1$ & Load0(99.20)  & Load1(98.33)  & Load2(97.92) & Load3(99.76) & 64 \\
    $$ & $3$ & Load0(99.61)  & Load1(100.00)  & Load2(100.00)  & Load3(100.00) & 77 \\
    $$ & $5$ & Load0(99.66)  & Load1(100.00)  & Load2(100.00)  & Load3(100.00) & 90 \\
    $$ & $10$ & Load0(99.75)  & Load1(100.00)  & Load2(100.00)  & Load3(100.00) & 132 \\
     $$ & $$ & $$ & $$ & $$ & $$ & $$ \\              

    8 & $1$ & Load0(99.72)  & Load1(100.00)  & Load2(99.41)  & Load3(99.75) & 65 \\
    $$ & $3$ & Load0(99.55)  & Load1(100.00)  & Load2(100.00)  & Load3(100.00) & 76 \\
    $$ & $5$ & Load0(99.65)  & Load1(100.00)  & Load2(100.00)  & Load3(100.00) & 91 \\
    $$ & $10$ & Load0(99.69)  & Load1(100.00)  & Load2(100.00)  & Load3(100.00) & 132 \\
    \bottomrule[1pt]
  \end{tabular}
\end{table}

\subsection{Results based on 2DPCA}
Having tested the performance of the 2DPCA based method with different dimension reduction, we determined the $d = 10$ defined in formula \ref{2dpcaeq3} as the best selection. The diagnostic results with 2DPCA for fault size being 0.014 and 0.021 are given in Table \ref{Result-0.014-2DPCA} and Table \ref{Result-0.021-2DPCA}. Taking Load0 as training samples, the classification rate could reach $100\%$ with $n=5$. When the sampling number of each training class is equal or greater than 5, most of the test cases could achieve a considerable classification rate in excess of 90\%. Unfortunately the two cases (marked with $\star$) are still around 75\%, and the reason is similar as discussed before.

\begin{table}[!hbp] \footnotesize
  \centering
  \renewcommand\arraystretch{1.1}               
  \renewcommand{\multirowsetup}{\centering}
  \setlength{\abovecaptionskip}{5pt}            
  \setlength{\belowcaptionskip}{0pt}            
  \caption{The classification rate based on 2DPCA with fault size being 0.014.}\label{Result-0.014-2DPCA}
    \begin{tabular}{ccccccc}
      \toprule[1pt]
      \multirow{2}{1cm}{\# of test} & \multirow{2}{0.4cm}{$n$\tnote{a}}
      & \multicolumn{5}{c}{Testing data} \\\cline{3-7}
      &  &  \multirow{1}{*}{Test1(\%)}  & \multirow{1}{*}{Test2(\%)} & \multirow{1}{*}{Test3(\%)}
      & \multirow{1}{*}{Test4(\%)} & \multirow{1}{*}{$T$(s)\tnote{b}} \\
      \hline
    1 & $1$ & Load0(99.97) & Load1(99.85) & Load2(99.41)  & Load3(94.50) & 51 \\
    $$ & $3$ & Load0(100.00)  & Load1(100.00)  & Load2(100.00)  & Load3(99.95) & 56 \\
    $$ & $5$ & Load0(100.00)  & Load1(100.00)  & Load2(100.00)  & Load3(100.00) & 63 \\
    $$ & $10$ & Load0(100.00)  & Load1(100.00)  & Load2(100.00)  & Load3(100.00) & 81 \\
     $$ & $$ & $$ & $$ & $$ & $$ & $$ \\              

    2 & $1$ & Load0(85.78)  & Load1(100.00)  & Load2(100.00)  & Load3$(75.45)^{\star}$ & 50 \\
    $$ & $3$ & Load0(88.67)  & Load1(100.00)  & Load2(100.00)  & Load3$(75.00)^{\star}$ & 56 \\
    $$ & $5$ & Load0(91.17)  & Load1(100.00)  & Load2(100.00)  & Load3$(76.46)^{\star}$ & 63 \\
    $$ & $10$ & Load0(92.56) & Load1(100.00)  & Load2(100.00)  & Load3$(78.26)^{\star}$ & 81 \\
     $$ & $$ & $$ & $$ & $$ & $$ & $$ \\              

    3 & $1$ & Load0(87.79)  & Load1(100.00)  & Load2(100.00)  & Load3$(77.69)^{\star}$ & 51 \\
    $$ & $3$ & Load0(91.56)  & Load1(99.96)  & Load2(100.00)  & Load3$(80.74)^{\star}$ & 58 \\
    $$ & $5$ & Load0(92.24)  & Load1(100.00)  & Load2(100.00)  & Load3$(80.66)^{\star}$ & 65 \\
    $$ & $10$ & Load0(93.45)  & Load1(100.00)  & Load2(100.00)  & Load3$(83.13)^{\star}$ & 84 \\
    $$ & $$ & $$ & $$ & $$ & $$ & $$ \\              

    4 & $1$ & Load0(99.25)  & Load1(86.17) & Load2(99.51)  & Load3(100.00) & 52 \\
    $$ & $3$ & Load0(99.55)  & Load1(85.94) & Load2(99.47)  & Load3(100.00) & 59 \\
    $$ & $5$ & Load0(99.65)  & Load1(84.63)  & Load2(99.26) & Load3(100.00) & 66 \\
    $$ & $10$ & Load0(99.89)  & Load1(83.64)  & Load2(99.53)  & Load3(100.00) & 85 \\
        \bottomrule[1pt]
  \end{tabular}
\end{table}

\begin{table}[!hbp] \footnotesize
  \centering
  \renewcommand\arraystretch{1.1}               
  \renewcommand{\multirowsetup}{\centering}
  \setlength{\abovecaptionskip}{5pt}            
  \setlength{\belowcaptionskip}{0pt}            
  \caption{The classification rate based on 2DPCA with fault size being 0.021.}\label{Result-0.021-2DPCA}
    \begin{tabular}{ccccccc}
      \toprule[1pt]
      \multirow{2}{1cm}{\# of test} & \multirow{2}{0.4cm}{$n$\tnote{a}}
      & \multicolumn{5}{c}{Testing data} \\\cline{3-7}
      &  &  \multirow{1}{*}{Test1(\%)}  & \multirow{1}{*}{Test2(\%)} & \multirow{1}{*}{Test3(\%)}
      & \multirow{1}{*}{Test4(\%)} & \multirow{1}{*}{$T$(s)\tnote{b}} \\
      \hline
    5 & $1$ & Load0(99.75)  & Load1(99.89)  & Load2(96.91)  & Load3(92.94) & 52 \\
    $$ & $3$ & Load0(99.75)  & Load1(100.00) & Load2(99.86)  & Load3(100.00) & 58 \\
    $$ & $5$ & Load0(100.00)  & Load1(100.00)  & Load2(100.00)  & Load3(99.04) & 65 \\
    $$ & $10$ & Load0(100.00)  & Load1(100.00)  & Load2(100.00)  & Load3(100.00) & 84 \\
    $$ & $$ & $$ & $$ & $$ & $$ & $$ \\              

   6 & $1$ & Load0(99.74)  & Load1(100.00)  & Load2(91.71)  & Load3(95.45) & 51 \\
    $$ & $3$ & Load0(99.75)  & Load1(100.00) & Load2(96.70)  & Load3(98.24) & 58 \\
    $$ & $5$ & Load0(99.75)  & Load1(100.00)  & Load2(96.08)  & Load3(99.95) & 65 \\
    $$ & $10$ & Load0(99.75)  & Load1(100.00)  & Load2(99.25)  & Load3(100.00) & 84 \\
    $$ & $$ & $$ & $$ & $$ & $$ & $$ \\              

    7 & $1$ & Load0(99.75)  & Load1(98.06)  & Load2(100.00)  & Load3(99.72) & 50 \\
    $$ & $3$ & Load0(99.75) & Load1(100.00)  & Load2(100.00)  & Load3(100.00) & 57 \\
    $$ & $5$ & Load0(99.75)  & Load1(100.00)  & Load2(100.00)  & Load3(100.00) & 63 \\
    $$ & $10$ & Load0(99.75) & Load1(100.00)  & Load2(100.00)  & Load3(100.00) & 82 \\
     $$ & $$ & $$ & $$ & $$ & $$ & $$ \\              

    8 & $1$ & Load0(99.75)  & Load1(100.00) & Load2(98.83) & Load3(100.00) & 51 \\
    $$ & $3$ & Load0(99.75) & Load1(100.00)  & Load2(100.00)  & Load3(100.00) & 58 \\
    $$ & $5$ & Load0(99.75)  & Load1(100.00)  & Load2(100.00)  & Load3(100.00) & 65 \\
    $$ & $10$ & Load0(99.75)  & Load1(100.00)  & Load2(100.00)  & Load3(100.00) & 83 \\
    \bottomrule[1pt]
  \end{tabular}
\end{table}

It is worth mentioning that the time consumption of 2DPCA is considerably smaller than PCA, especially when $n$ is larger. The detailed comparison is given in Table \ref{Result-T-Diversity}. It is clear that 2DPCA method is more efficient than PCA method when using FFT spectrum images for fault diagnosis of bearings.

\begin{table}[!hbp] \footnotesize
  \centering
  \renewcommand\arraystretch{1.1}               
  \renewcommand{\multirowsetup}{\centering}
  \setlength{\abovecaptionskip}{5pt}            
  \setlength{\belowcaptionskip}{0pt}            
  \caption{The time consumption diversity of Load0 as training with fault size being 0.014.}\label{Result-T-Diversity}
  \begin{threeparttable}
    \begin{tabular}{cccccc}
      \toprule[1pt]
      \multirow{1}{2cm}{Training data} & \multirow{1}{2cm}{Testing data} & \multirow{1}{1cm}{$n$}
      & \multirow{1}{1.6cm}{$T_{pca}$(s)} & \multirow{1}{1.6cm}{$T_{2dpca}$(s)} & \multirow{1}{1.6cm}{$\triangle T$(s)\tnote{*}} \\
      \hline

      Load0 & Load0 & 1 & 68 & 51 & 17 \\
      & & 3 & 82 & 58 & 24 \\
      & & 5 & 96 & 64 & 32 \\
      & & 10 & 144 & 82 & 62 \\
      $$ & $$ & $$ & $$ & $$ & $$ \\

      & Load1 & 1 & 71 & 52 & 19 \\
      & & 3 & 85 & 59 & 26 \\
      & & 5 & 95 & 65 & 30 \\
      & & 10 & 145 & 83  & 62 \\
      $$ & $$ & $$ & $$ & $$ & $$ \\
      & Load2 & 1 & 65 & 51 & 14 \\
      & & 3 & 80 & 57 & 23 \\
      & & 5 & 99 & 64 & 35 \\
      & & 10 & 143 & 82 & 61 \\
      $$ & $$ & $$ & $$ & $$ & $$ \\
      & Load3 & 1 & 66 & 51 & 15 \\
      & & 3 & 80 & 56 & 24 \\
      & & 5 & 95 & 63 & 32 \\
      & & 10 & 145 & 81 & 64 \\
      \bottomrule[1pt]
    \end{tabular}
    \begin{tablenotes}
    \footnotesize
      \item[*] $\triangle T = T_{pca}-T_{2dpca}$, that is: the time consumption difference between PCA and 2DPCA.
    \end{tablenotes}
  \end{threeparttable}
\end{table}

\subsection{Discussion}
The spectrum image of a given vibration signal is constructed based on FFT. In fact the spectrum is just a vector of amplitudes. Nevertheless it is not easy to mine useful knowledge directly from the vector. The spectrum image is a different view of the vector and provides a new way to dig information. We also carry out the similar tests by taking FFT spectrum amplitudes as the features, where PCA and the same minimum distance method are used for fault classification. Similarly, after testing the performance of different dimension reduction with PCA and selecting the best case, contribution of selected components with 90\% is also designated in this section.  

The results are shown in Table \ref{Result-0.014-FFT} and Table \ref{Result-0.021-FFT}. Obviously the classification performance using FFT amplitude as features is inferior to that of using the spectrum images, especially when the testing load condition is different from the training load condition.

\begin{table} [!hbp] \footnotesize
  \centering
  \renewcommand\arraystretch{1.1}               
  \renewcommand{\multirowsetup}{\centering}
  \setlength{\abovecaptionskip}{5pt}            
  \setlength{\belowcaptionskip}{0pt}            
  \caption{The classification rate with FFT amplitude with fault size being 0.014.}\label{Result-0.014-FFT}
    \begin{tabular}{ccccccc}
      \toprule[1pt]
      \multirow{2}{1cm}{\# of test} & \multirow{2}{0.4cm}{$n$\tnote{a}}
      & \multicolumn{5}{c}{Testing data} \\\cline{3-7}
      &  &  \multirow{1}{*}{Test1(\%)}  & \multirow{1}{*}{Test2(\%)} & \multirow{1}{*}{Test3(\%)}
      & \multirow{1}{*}{Test4(\%)} & \multirow{1}{*}{$T$(s)\tnote{b}} \\
      \hline
    1 & $1$ & Load0(85.51) & Load1(62.81)  & Load2(67.84)  & Load3(66.20) & 10 \\
    $$ & $3$ & Load0(93.41)  & Load1(63.69)  & Load2(59.66)  & Load3(61.17) & 10 \\
    $$ & $5$ & Load0(96.78)  & Load1(57.50)  & Load2(54.91) & Load3(51.51) & 10 \\
    $$ & $10$ & Load0(99.39)  & Load1(51.25)  & Load2(53.75)  & Load3(50.00) & 11 \\
    $$ & $$ & $$ & $$ & $$ & $$ & $$ \\              

    2 & $1$ & Load0(73.64)  & Load1(99.09)  & Load2(73.81)  & Load3(73.04) & 9 \\
    $$ & $3$ & Load0(76.31)  & Load1(99.89)  & Load2(75.00)  & Load3(75.06) & 9 \\
    $$ & $5$ & Load0(75.00)  & Load1(100.00)  & Load2(75.00)  & Load3(75.00) & 9 \\
    $$ & $10$ & Load0(75.00)  & Load1(100.00)  & Load2(75.00)  & Load3(75.00) & 11 \\
    $$ & $$ & $$ & $$ & $$ & $$ & $$ \\              

    3 & $1$ & Load0(77.81)  & Load1(74.99) & Load2(90.99) & Load3(76.25) & 9 \\
    $$ & $3$ & Load0(78.59) & Load1(74.17)  & Load2(98.47)  & Load3(77.01) & 9 \\
    $$ & $5$ & Load0(76.65)  & Load1(75.00)  & Load2(99.91) &  Load3(75.00) & 10 \\
    $$ & $10$ & Load0(75.66)  & Load1(75.00)  & Load2(100.00)  & Load3(75.00) & 12 \\
    $$ & $$ & $$ & $$ & $$ & $$ & $$ \\              

    4 & $1$ & Load0(73.56)  & Load1(75.00)  & Load2(73.40)  & Load3(97.61) & 9 \\
    $$ & $3$ & Load0(74.86)  & Load1(75.00)  & Load2(75.00) & Load3(98.94) & 9 \\
    $$ & $5$ & Load0(75.00)  & Load1(75.00)  & Load2(75.00)  & Load3(99.01) & 10 \\
    $$ & $10$ & Load0(75.00)  & Load1(75.00)  & Load2(75.00)  & Load3(99.31) & 11 \\
    \bottomrule[1pt]
  \end{tabular}
\end{table}

\begin{table} [!hbp] \footnotesize
  \centering
  \renewcommand\arraystretch{1.1}               
  \renewcommand{\multirowsetup}{\centering}
  \setlength{\abovecaptionskip}{5pt}            
  \setlength{\belowcaptionskip}{0pt}            
  \caption{The classification rate with FFT amplitude with fault size being 0.021.}\label{Result-0.021-FFT}
    \begin{tabular}{ccccccc}
      \toprule[1pt]
      \multirow{2}{1cm}{\# of test} & \multirow{2}{0.4cm}{$n$\tnote{a}}
      & \multicolumn{5}{c}{Testing data} \\\cline{3-7}
      &  &  \multirow{1}{*}{Test1(\%)}  & \multirow{1}{*}{Test2(\%)} & \multirow{1}{*}{Test3(\%)}
      & \multirow{1}{*}{Test4(\%)} & \multirow{1}{*}{$T$(s)\tnote{b}} \\
      \hline
    5 & $1$ & Load0(85.08)  & Load1(58.45)  & Load2(59.14)  & Load3(60.05) & 9 \\
    $$ & $3$ & Load0(95.58)  & Load1(65.08)  & Load2(68.80)  & Load3(63.63) & 9 \\
    $$ & $5$ & Load0(98.96)  & Load1(66.78)  & Load2(69.65)  & Load3(65.17) & 9 \\
    $$ & $10$ & Load0(99.76)  & Load1(70.14)  & Load2(72.04)  & Load3(65.30) & 10 \\
     $$ & $$ & $$ & $$ & $$ & $$ & $$ \\              

    6 & $1$ & Load0(66.04)  & Load1(86.88) & Load2(70.35)  & Load3(60.31) & 8 \\
    $$ & $3$ & Load0(72.36)  & Load1(97.33)  & Load2(72.59)  & Load3(67.45) & 9 \\
    $$ & $5$ & Load0(74.40)  & Load1(99.70)  & Load2(72.86)  & Load3(69.40) & 9 \\
    $$ & $10$ & Load0(74.81)  & Load1(99.99)  & Load2(73.53) & Load3(70.94) & 10 \\
     $$ & $$ & $$ & $$ & $$ & $$ & $$ \\              

    7 & $1$ & Load0(71.17)  & Load1(69.04) & Load2(94.28)  & Load3(64.80) & 8 \\
    $$ & $3$ & Load0(74.74)  & Load1(73.97)  & Load2(99.99)  & Load3(69.79) & 9 \\
    $$ & $5$ & Load0(74.97) & Load1(74.95)  & Load2(99.99)  & Load3(71.80) & 9 \\
    $$ & $10$ & Load0(75.08)  & Load1(75.00) & Load2(100.00)  & Load3(74.17) & 10 \\
    $$ & $$ & $$ & $$ & $$ & $$ & $$ \\              

    8 & $1$ & Load0(73.40)  & Load1(68.90)& Load2(71.17)  & Load3(96.39) & 9 \\
    $$ & $3$ & Load0(75.22)  & Load1(75.15)  & Load2(73.70)  & Load3(96.99) & 9 \\
    $$ & $5$ & Load0(75.50)  & Load1(75.05)  & Load2(75.42)  & Load3(97.33) & 9 \\
    $$ & $10$ & Load0(76.33)  & Load1(75.00)  & Load2(75.89)  & Load3(99.70) & 10 \\
    \bottomrule[1pt]
  \end{tabular}
\end{table}

Some other remarks are as follows.
\begin{enumerate}[(1)]
  \item By comparing Table \ref{Result-0.014-PCA}, Table \ref{Result-0.021-PCA} with Table \ref{Result-0.014-2DPCA} and Table \ref{Result-0.021-2DPCA}, it can be found that bearing fault diagnosis based on 2DPCA can achieve better performance than PCA in most cases.
  \item Time consumption of processing based on 2DPCA is much less than PCA, especially when the number of training samples per class is large.
  \item In general, a larger $n$ could obtain a higher classification rate (see Table \ref{Result-0.014-PCA}, Table \ref{Result-0.021-PCA}, Table \ref{Result-0.014-2DPCA} and Table \ref{Result-0.021-2DPCA})). When using the spectrum image as the feature, an acceptable classification rate can still be achieved with only one single training image.
\end{enumerate}

In order to illustrate the potential application of proposed method in bearing fault diagnosis, a comparative study between the present work and published literature is presented in Table \ref{tab4}. Adopting the same faulty bearing data collected from the Case Western Reserve University \cite{CWRUBDC-Bearings}, most of the previous works considered only single load condition, where the training load condition and the testing load condition are the same. Only a few works evaluated the vibration data of multiple loading conditions. As shown in Table \ref{tab4}, bearing fault diagnosis based on SLLEP under load 3hp condition, was carried out to classify bearings with fault size being 0.021in using minimum-distance classifier in \cite{li2011supervised}. In \cite{yang2007}, with FDF as feature, SVMs and fractal dimension were employed to diagnose the bearings with fault size being 0.014in and 0.021in. In \cite{Liu2014}, feature extraction based on LMD analysis method and MSE was put forward to perform fault diagnosis of bearings under load 3hp condition. In \cite{Li2016}, bearings under load 2hp condition, were conducted fault diagnosis based on MPE and ISVM-BT. Moreover taken all load conditions into account, the improved distance evaluation technique and ANFIS were also employed to diagnose bearings with seven health cases in \cite{lei2008}.

However the classification rate of the proposed method can achieve 100\% in case the training and testing load condition are same as shown in Table \ref{Result-0.014-2DPCA} and Table \ref{Result-0.021-2DPCA}. And the classification rate is still high when the training and testing load condition are different.
\begin{table}
\begin{center}
\caption{Comparisons between the current work and some published work.}  \label{tab4}
\begin{tabular}{ccccc}
\hline
References &   Load conditions & No. of  & Classification \\
 &    & training samples & rate\\
\hline
Li et al. \cite{li2011supervised} &   Single  & 100  & 98.33\% \\
Yang et al. \cite{yang2007} &   Single  & 118  & 95.253 \% (0.014in)\\
 &     &   & 99.368 \% (0.021in)\\
Liu et al. \cite{Liu2014} &   Single  &  15 & 100\%\\
Li et al. \cite{Li2016} &   Single  &  80 & 100\%\\
Lei et al. \cite{lei2008} &   Multiple & 20 & 91.42\% \\
The proposed method &   Multiple & 10 & 95.65\% (0.014in) \\
                    &    &  & 99.90\%(0.021in) \\
 \hline
\end{tabular}
\end{center}
\end{table}

\section{Conclusion}
In this paper, the spectrum images were proposed as the features for fault diagnosis of bearings. The spectrum images of vibration signals could be simply obtained through FFT. After processed with 2DPCA, the corresponding eigen images were extracted. The classification of faults was realized with a simple minimum distance criterion based on the eigen images. The effectiveness of the proposed method was demonstrated with experimental vibration signals. As a different view of FFT spectrum, the images could significantly improve the performance of fault diagnosis. When the training sample is very limited, e.g. only one training image, the proposed method can still achieve high accuracy.

\section*{Acknowledgements}
The work was supported by National Natural Science Foundation of China (51475455), Natural Science Foundation of Jiangsu (BK20141127), the Fundamental Research Funds for the Central Universities (2014Y05), and the project funded by the Priority Academic Program Development of Jiangsu Higher Education Institutions (PAPD).

\bibliography{mst_ver2}

\end{document}